\documentclass{llncs}

\usepackage[pdftex]{graphicx}
\usepackage{amsmath}
\usepackage{multirow}
\usepackage{rotating}

\usepackage{longtable}

\begin{document}

\title{Early Experiences with Crowdsourcing\\Airway Annotations in Chest CT}

\author{Veronika Cheplygina$^{1,2}$, Adria Perez-Rovira$^{1,3}$,
Wieying Kuo$^{3,4}$,\\ Harm A. W. M. Tiddens$^{3,4}$, Marleen de Bruijne$^{1,5}$}

\institute{$^1$ Biomedical Imaging Group Rotterdam, Depts Medical Informatics and Radiology, Erasmus Medical Center, The Netherlands\\
$^2$ Pattern Recognition Laboratory, Delft University of Technology, The Netherlands\\
$^3$ Dept of Pediatric Pulmonology and Allergology, Erasmus Medical Center - Sophia Children's Hospital, Rotterdam, The Netherlands\\
$^4$ Dept of Radiology, Erasmus Medical Center, Rotterdam, The Netherlands\\
$^5$ Image Section, Dept of Computer Science, University of Copenhagen, Denmark\\
}

\maketitle

\begin{abstract}
Measuring airways in chest computed tomography (CT) images is important for characterizing diseases such as cystic fibrosis, yet very time-consuming to perform manually. Machine learning algorithms offer an alternative, but need large sets of annotated data to perform well. We investigate whether crowdsourcing can be used to gather airway annotations which can serve directly for measuring the airways, or as training data for the algorithms. We generate image slices at known locations of airways and request untrained crowd workers to outline the airway lumen and airway wall. Our results show that the workers are able to interpret the images, but that the instructions are too complex, leading to many unusable annotations. After excluding unusable annotations, quantitative results show medium to high correlations with expert measurements of the airways. Based on this positive experience, we describe a number of further research directions and provide insight into the challenges of crowdsourcing in medical images from the perspective of first-time users.
\end{abstract}

\section{Introduction}

Respiratory diseases are a major cause of death and disability and are responsible for three out of the top five causes of death worldwide~\cite{factsheet}. Chest computed tomography (CT) is an important tool to characterize and monitor lung diseases. Quantification of structural abnormalities in the lungs, such as bronchiectasis, air trapping and emphysema, is needed to track disease progression or to predict patient outcomes. We have recently shown that, the airway-to-vessel ratio (AVR) is an objective measurement of bronchiectasis which is sensitive to detect early lung disease~\cite{tiddens2010cystic,mott2013assessment}. Unfortunately, manual measurements of the airways and adjoining arteries suffer from intra- and inter-observer variation and are very time-consuming (8-16 hours per chest CT).

Computer algorithms can be used to improve accuracy and efficiency of the measurements. The first step is to extract the airways and vessels from the scan. Machine learning techniques learn from example images which have been manually annotated, and have shown to be very effective for such extraction tasks~\cite{lo2010vessel}.  However, these techniques require a large amount of annotated images, which is also expensive and time-consuming.

We therefore propose to use the wisdom of the crowd to gather annotations. In crowdsourcing, untrained internet users (knowledge workers or KWs) carry out human intelligence tasks (HITs), such as annotating images. The KWs are unpaid volunteers, or receive a small financial reward for each task. Early research into crowdsourcing for medical images~\cite{maier2014crowdsourcing,nguyen2012distributed,mitry2015crowdsourcing,maier2015crowdtruth} showed that non-expert workers were able to carry out a range of HITs relatively well; our goal is to investigate whether this is true for airway measurement in chest CT.

In this paper we describe our early experiences with crowdsourcing airway measurements in chest CT images. In Section~\ref{sec:methods} we describe how we generate 2D slices, how we collect annotations from the KWs and how the annotations are processed. Section~\ref{sec:experiments} describes the data and the number of annotations collected, followed by a presentation of the results in Section~\ref{sec:results}. We discuss our findings and steps for future research in Section~\ref{sec:discussion}, followed by a conclusion in Section~\ref{sec:conclusion}.

\section{Methods}\label{sec:methods}

Our main question for this study was whether non-expert workers would be able to annotate airways in chest CT images. By ``an airway annotation'' we understand two outlines: one of the airway lumen (inner airway) and one of the airway wall (outer airway).  Annotating an airway consists of two steps: localizing an airway, and creating the outlines. In this study we focused on the second question only. We therefore acquired annotations using already existing 3D voxel coordinates and orientations as a starting point.

We used 3D voxel coordinates, at which experts have previously annotated airways using the Myrian\textsuperscript{TM} software. As we could not reproduce how the software determines the orientations, we used an airway segmentation algorithm for this step. The method starts with an initial volumetric segmentation of the airways, rescales it isotropically and uses front propagation to obtain airway centerlines~\cite{petersen2014optimal}, which give us the orientations.

\begin{figure}
\includegraphics[width=0.7\textwidth]{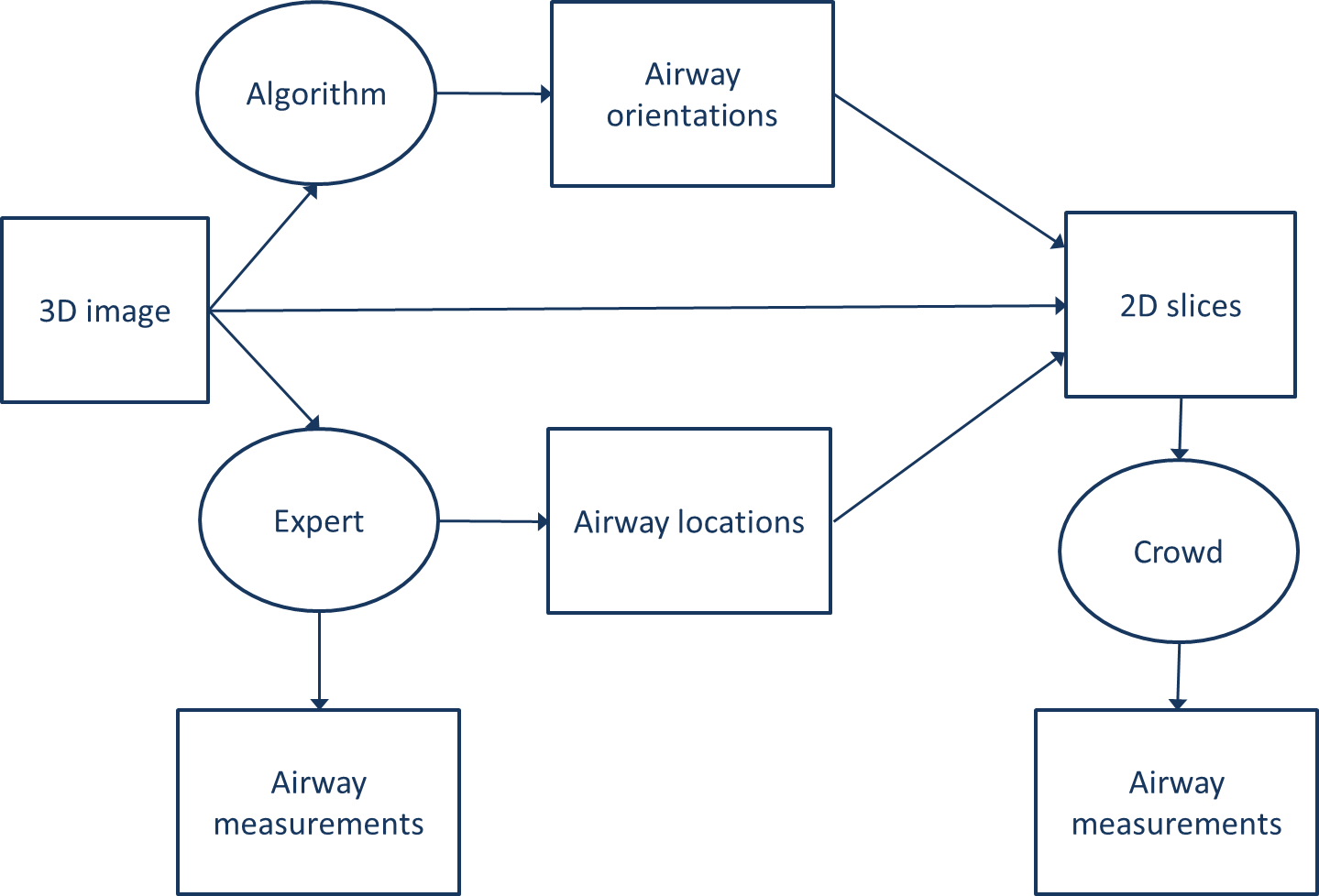}
\caption{Overview of the method. A 3D image is annotated by experts. The locations and orientations of the annotations are then used to generate 2D slices of the image, which are then annotated by the workers.}
\label{fig:overview}
\end{figure}

Using the 3D coordinates and orientations, we generate 2D slices (described in more detail in Section~\ref{sec:generation}), which are annotated by the KWs. This allows for a comparison of airway measurements between the experts and the KWs. Fig.~\ref{fig:overview} shows a global overview of our method.

\subsection{Image Generation}~\label{sec:generation}

\begin{figure}[ht]
\includegraphics[width=0.9\textwidth]{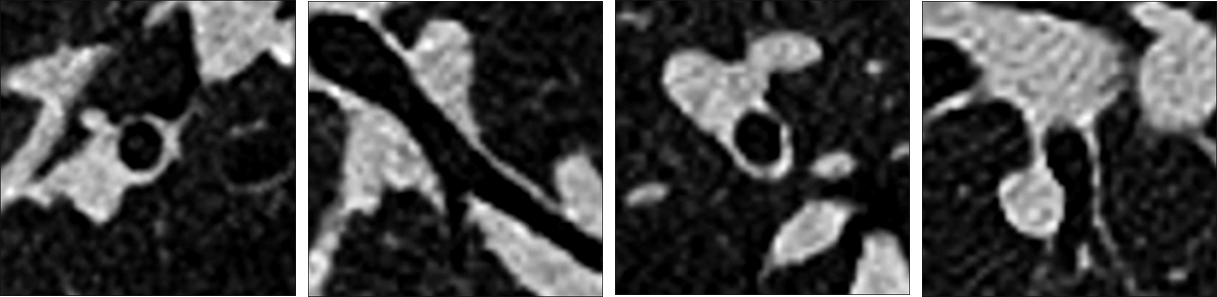}
\caption{Slices of $50\times50$ voxels showing four views of an airway, from left to right: original orientation, saggital, coronal and axial views. An airway cross-section appears as a dark circle (airway lumen) with a light ring (airway wall) around it}
\label{fig:orientations}
\end{figure}

Given a 3D location and an orientation vector, we generated a slice of $50\times50$ voxels, perpendicular to that orientation. Because of possible segmentation errors, an airway was not always visible. We therefore also generated slices in axial, coronal and saggital views, in total generating four different images per airway.
We used cubic interpolation and an intensity range between -950 and 550 Hounsfield units for better contrast, as recommended by the experts. An example is shown in Fig.~\ref{fig:orientations}.

\subsection{Annotation Software}

Amazon Mechanical Turk or MTurk~\cite{chen2011opportunities} is an internet-based crowdsourcing platform that allows untrained internet users, known as knowledge workers (KWs) to perform tasks, known as human intelligence tasks (HITs), for a small (in the order of \$0.05) financial reward. We integrated our annotation interface into MTurk by supplying a dynamic webpage, built with HTML5 and Javascript. The interface originally contained a freehand tool for creating annotations, which was later replaced by an ellipse tool, which more closely resembled the tool used by the experts.

\begin{figure}[ht]
\includegraphics[width=0.98\textwidth]{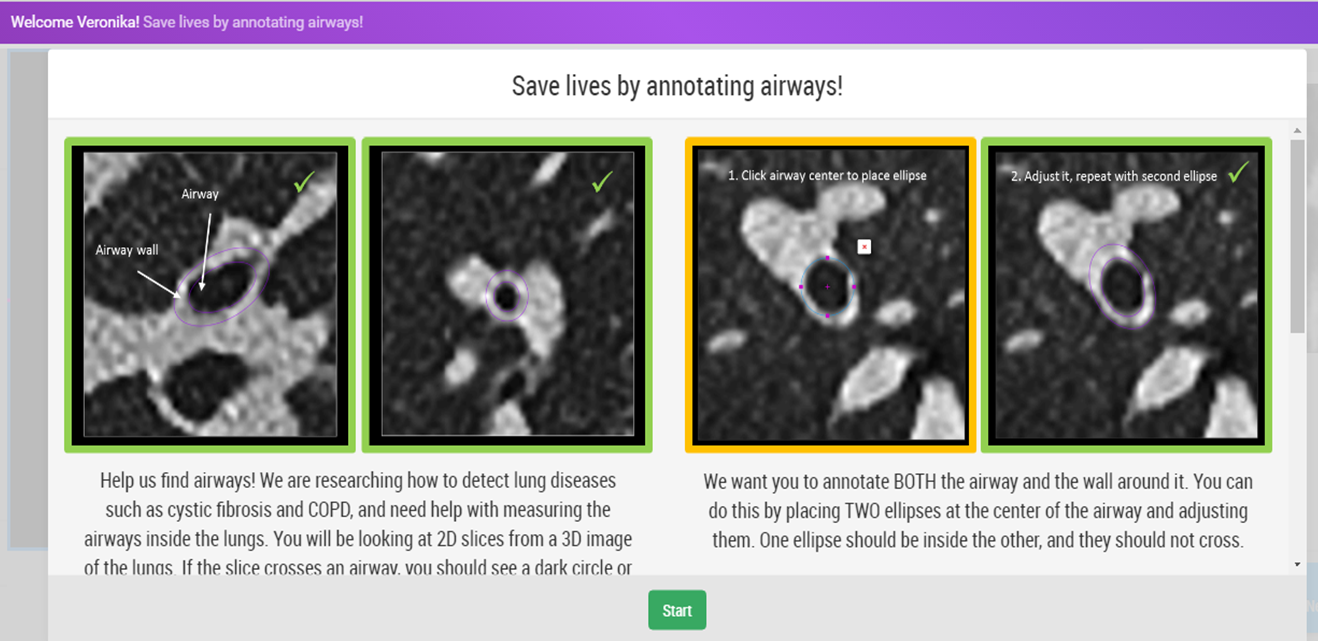}

\caption{Screenshot of the instructions given to the KWs for the task with the ellipse tool. The scrollbar on the right shows that there are more instructions than are visible in one screen.}
\label{fig:instructions}
\end{figure}

The details of our HIT, which the KWs could see when searching for HITs, are shown in Table~\ref{tab:hit}, and a screenshot of the instructions is shown in Fig.~\ref{fig:instructions}. The KWs were instructed to draw two ellipses outlining the airway lumen and the airway wall, or to draw a small circle in the corner of the image, if no airway is visible. For each HIT, the software recorded an anonymized ID of the KW and the coordinates of the annotations.

\begin{table}[ht]
\begin{tabular}{p{2cm}  p{10cm}}
Title & Save lives by annotating airways! \\
Description & Draw two contours to annotate an airway (dark circle or ellipse) in image from a lung scan \\
Keywords &
image, annotation, contour, draw, drawing, segmentation, medical \\
\end{tabular}
\caption{Details of HIT on Mechanical Turk}
\label{tab:hit}
\end{table}

\subsection{Airway Measurement}~\label{sec:measurement}

We applied a simple filtering step to discard unusable annotations. The following annotations are discarded:
\begin{itemize}
    \item no ellipses
    \item an odd number of ellipses
    \item an even number of ellipses, but the distance between centers of paired ellipses (pairs were assigned based on center distance) is larger than 10 voxels
\end{itemize}

For the remaining usable annotations, we measured the areas of the inner and outer ellipse, in order to compare them to the expert annotations. We perform the comparisons for each KW annotation individually, as well as for a combined measurement of the KWs. To obtain the combined measurements, we used only images with at least three usable annotations, and took the median of the areas.

\section{Experiments}\label{sec:experiments}

\subsection{Data}
For this preliminary experiment we used 1 inspiratory pediatric CT scan from a cohort of 24 subjects from a study~\cite{perez2015automated,kuo2015assessment}, collected at the Erasmus MC - Sophia Children's Hospital. In this scan, 76 airways were annotated by an expert using Myrian software. The expert localized an airway, outlined the inner and outer airway, and recorded the measurements of the areas.

\subsection{Crowd Annotations}

We generated a total of $76\times4=308$ images using the method described in Section~\ref{sec:generation}. We randomly created HITs with 10 images per HIT. A KW could request a HIT, annotate 10 images, and then submit the HIT. The KWs were paid \$0.10 per completed HIT. Only KWs who had previously done at least 100 HITs with an acceptance rate of 90\% could request the HITs.

We first collected 1 annotation per image with freehand tool. As we will describe in Section~\ref{sec:results}, it became clear that an ellipse tool was needed. With the ellipse tool, we collected 10 annotations per image.

\section{Results}\label{sec:results}


\subsection{Annotations}
\begin{figure}[ht]
\includegraphics[width=0.98\textwidth]{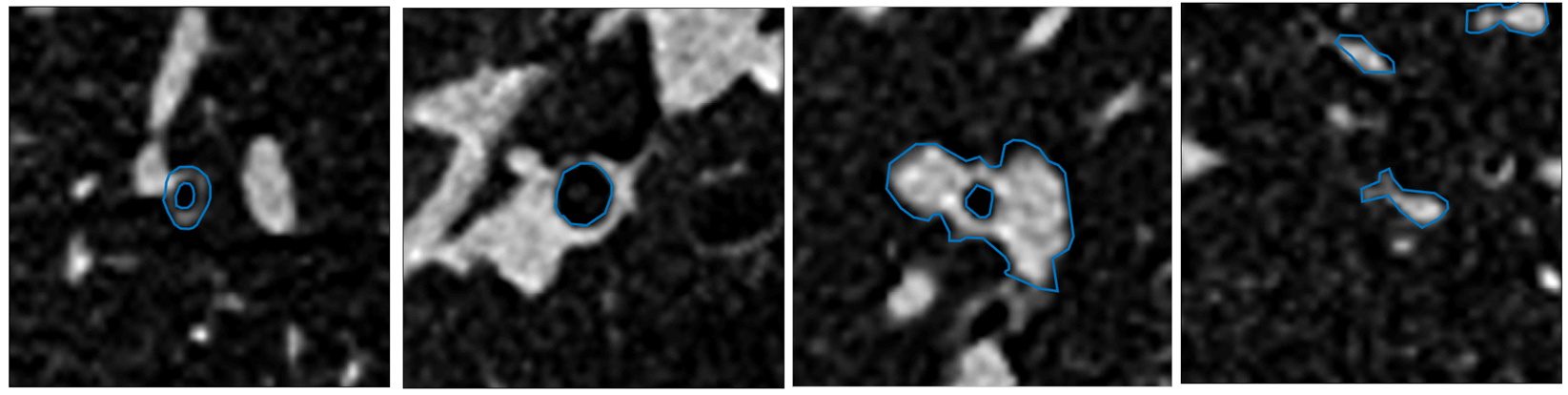}

\includegraphics[width=0.98\textwidth]{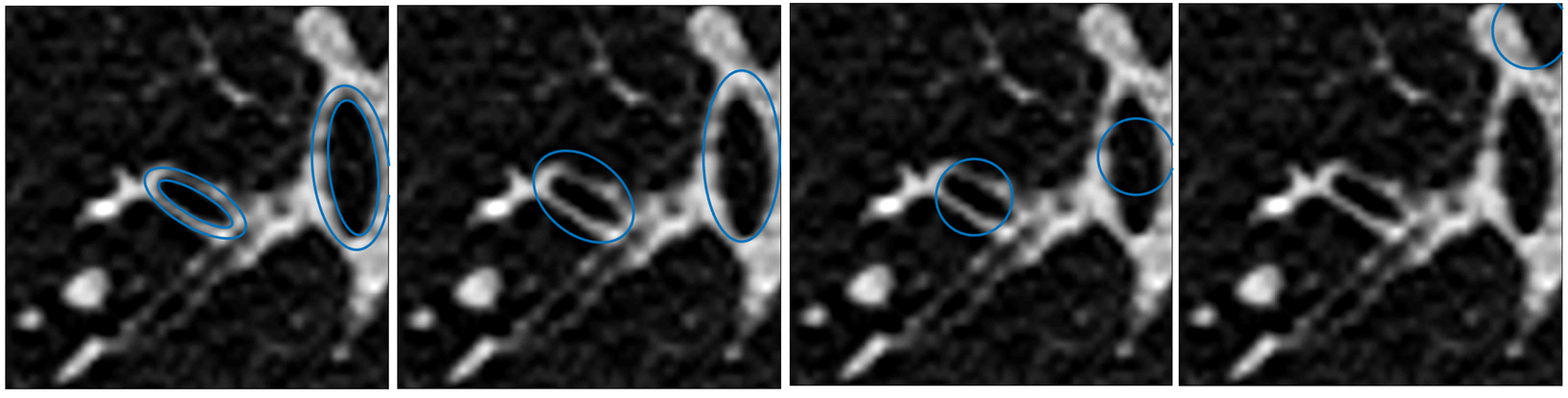}

\caption{Top: Annotations acquired with the freehand tool for four different images: correct annotation and three incorrect annotations. Bottom: Annotations acquired with the ellipse for the same image: correct annotations and three incorrect annotations.}
\label{fig:annot}
\end{figure}

We first collected 1 annotation per image with the freehand tool. A selection of the results is shown in Fig.~\ref{fig:annot} (top). Most of the workers attempted to annotate something in the image (i.e., were not spammers), but many annotations were not usable. For example, many workers misunderstood the instructions, annotated vessels instead of airways, drew only one contour or drew non-ellipsoidal contours. We concluded that this tool allowed too many degrees of freedom, and opted for the more controlled ellipse tool.

With the ellipse tool, we collected 10 annotations per image. However, based on our experience with the freehand tool, to reduce costs we did not gather annotations for all the images. In the end, with the ellipse tool 90 of the 308 images were annotated, resulting in 900 annotations.

A selection of the results with the ellipse tool in shown in Fig.~\ref{fig:annot} (bottom). Using the tool eliminated the problem of non-ellipsoidal airways. However, the problems of either a single contour, or workers annotating vessels, were still present. While the annotations still were not perfect, we decided to do proceed with an initial analysis of the annotations.

\subsection{Airway Measurement}

We filtered unusable annotations as described in Section~\ref{sec:measurement}. Out of 900 annotations, 610 were found to be unusable. Of these 610, 133 annotations contained no ellipse, and 445 annotations contained only a single ellipse. For annotations with a single ellipse, there are three possible causes: spam, the worker indicating ``no airway visible'', or the worker misunderstood the instructions. To better differentiate between these causes, we looked at whether the ellipse was adjusted, indicating that the worker tried to annotate something. This was the case for 244 of the 445 annotations with a single ellipse. Although we do not analyse these annotations in this preliminary study, we note that these annotations still could be used to measure airways.

Next we focus on the the 290 usable annotations, i.e. where the worker placed ellipses in pairs. Of these, 256 annotations contained a single pair, 25 annotations contained two pairs, and a further 6 annotations contained three pairs. For this preliminary study, we only consider the annotations with a single pair for further analysis.

\begin{figure}[ht]
\includegraphics[width=0.45\textwidth]{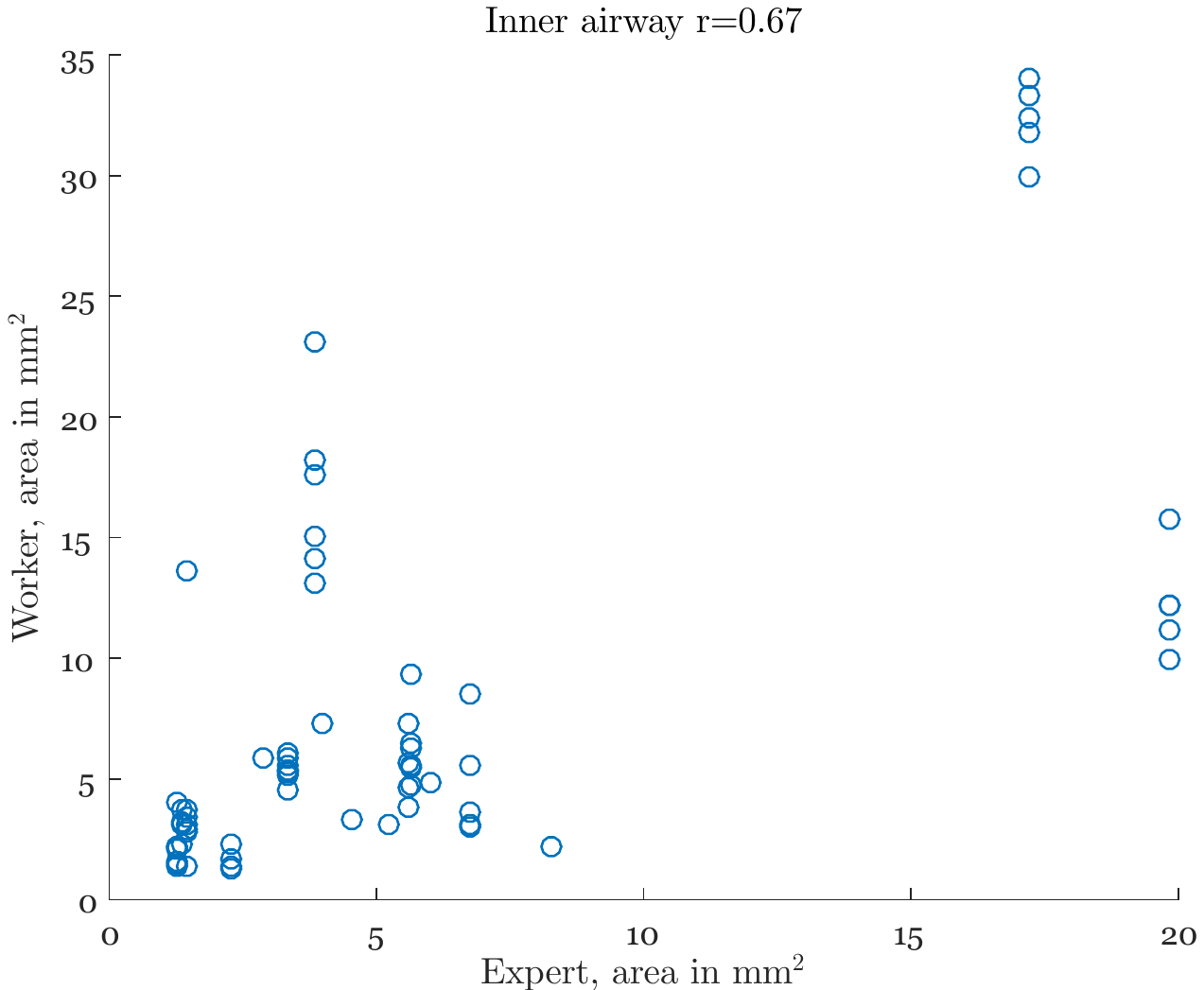}
\includegraphics[width=0.45\textwidth]{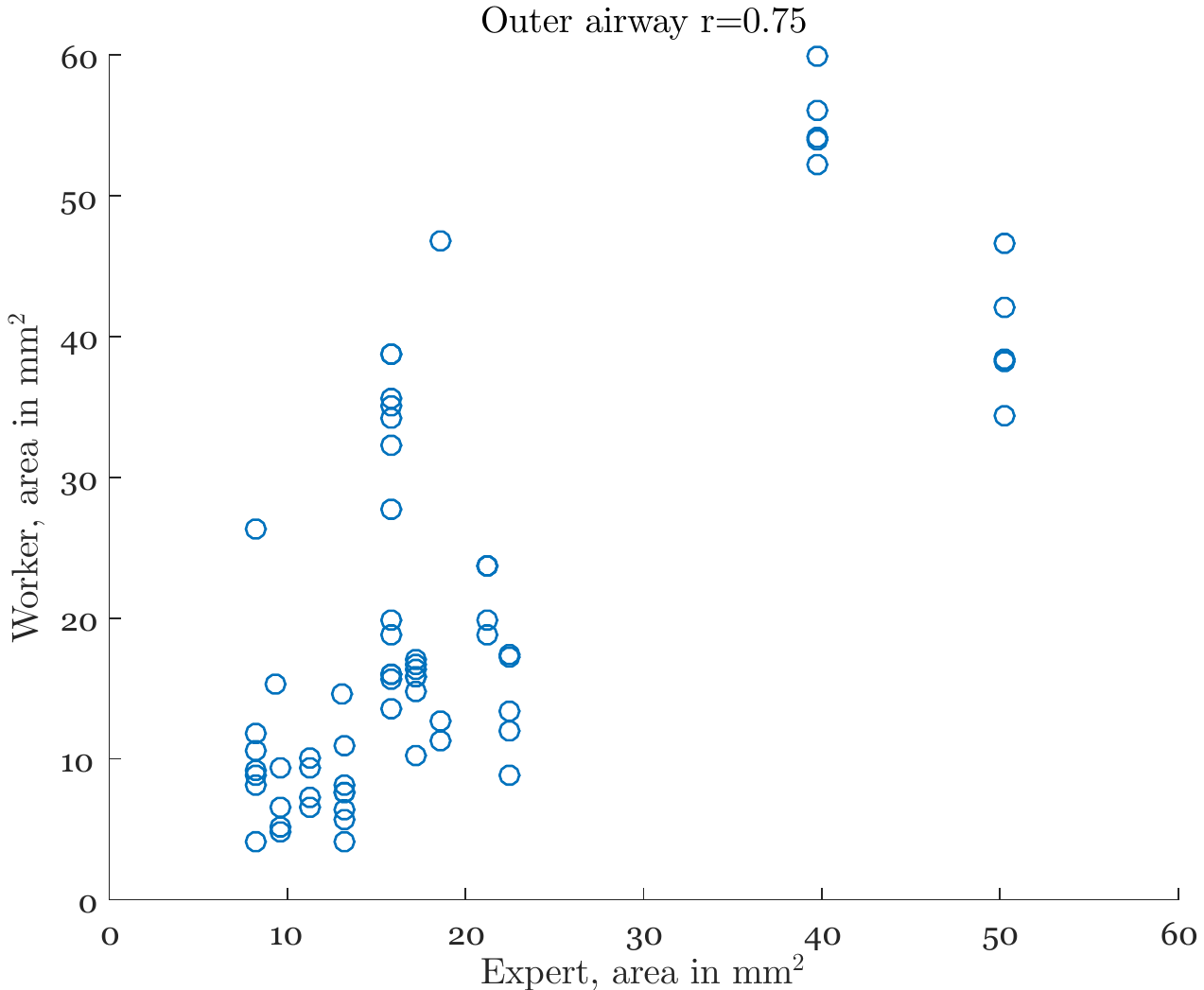}

\includegraphics[width=0.45\textwidth]{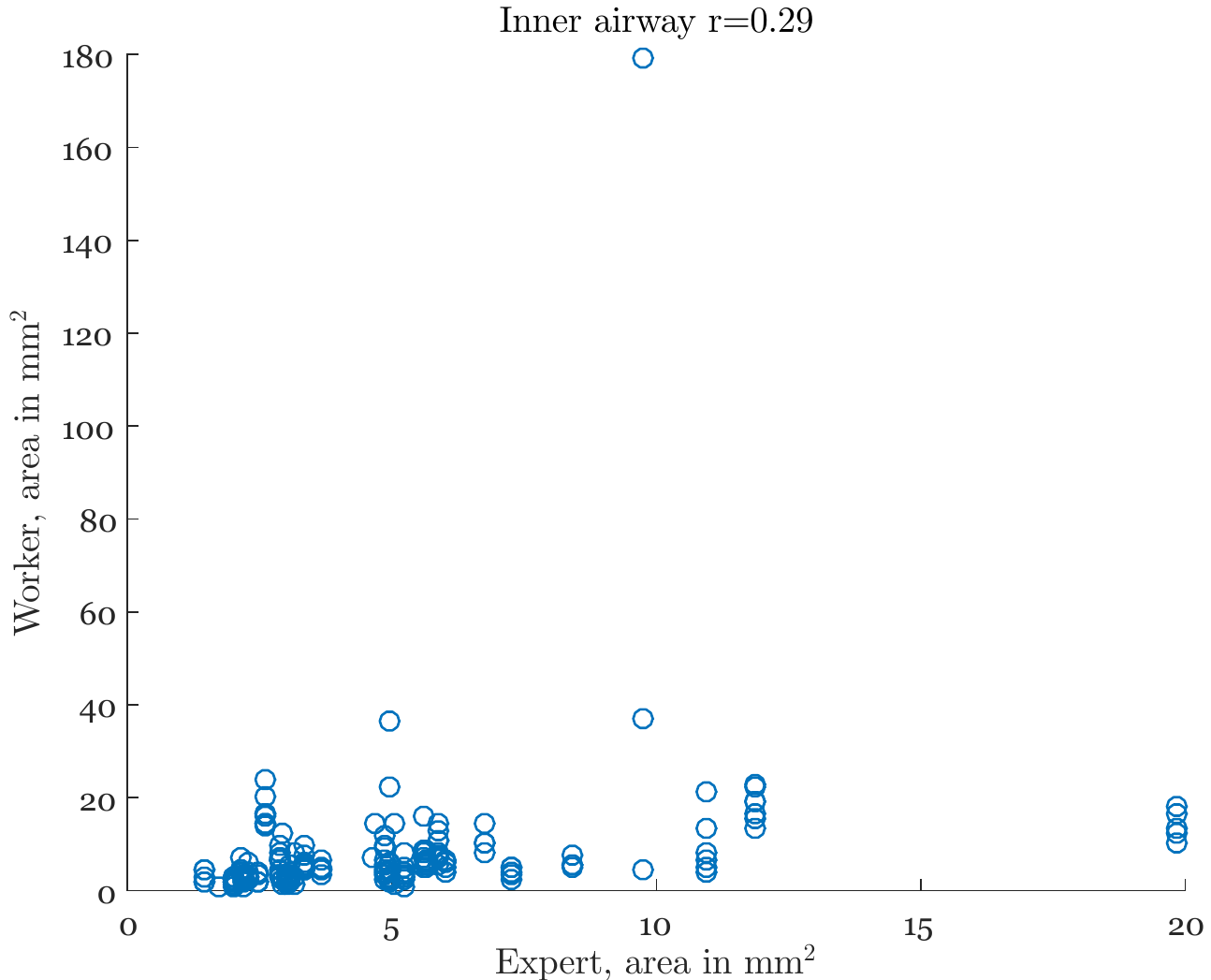}
\includegraphics[width=0.45\textwidth]{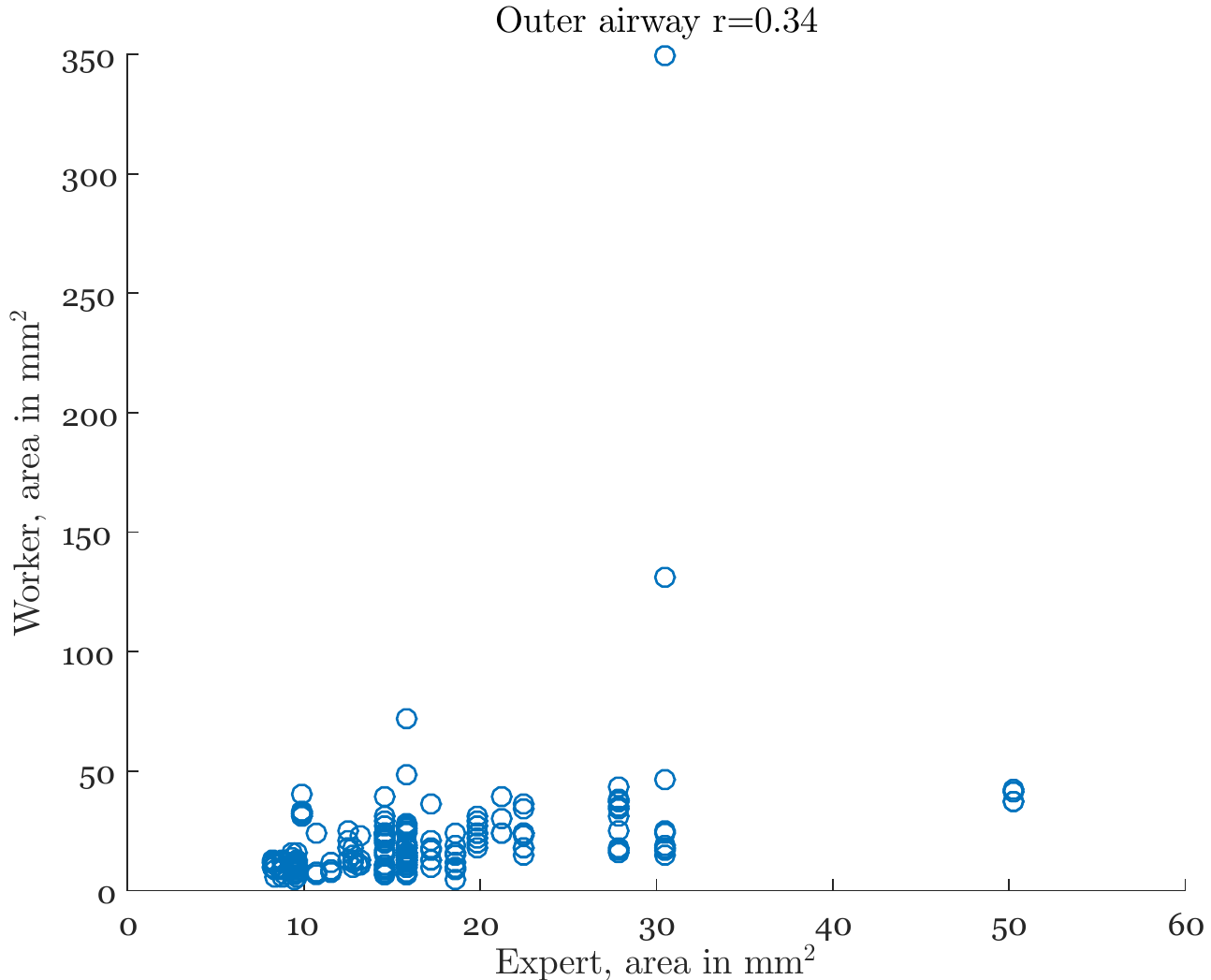}
\caption{Scatter plots of expert vs individual worker measurements of the areas. Top: original orientation, bottom: axes-parallel orientations. Left: airway lumen, right: airway wall. $r$ indicates Pearson's correlation.}
\label{fig:individual}
\end{figure}

To assess correctness of the annotations, we create expert-vs-worker plots of two quantities: area of the airway lumen and area of the airway wall. We show the annotations for the original orientation in Fig.~\ref{fig:individual} (top), and the annotations for the saggital, coronal and axial orientations in Fig.~\ref{fig:individual} (bottom). The correlations for the original orientations are medium to high, although workers tend to overestimate the airway lumen. The correlations for the other orientations are, understandably, weaker. Possibly here workers annotate other structures that are visible in the images.

\begin{figure}[ht]
\includegraphics[width=0.45\textwidth]{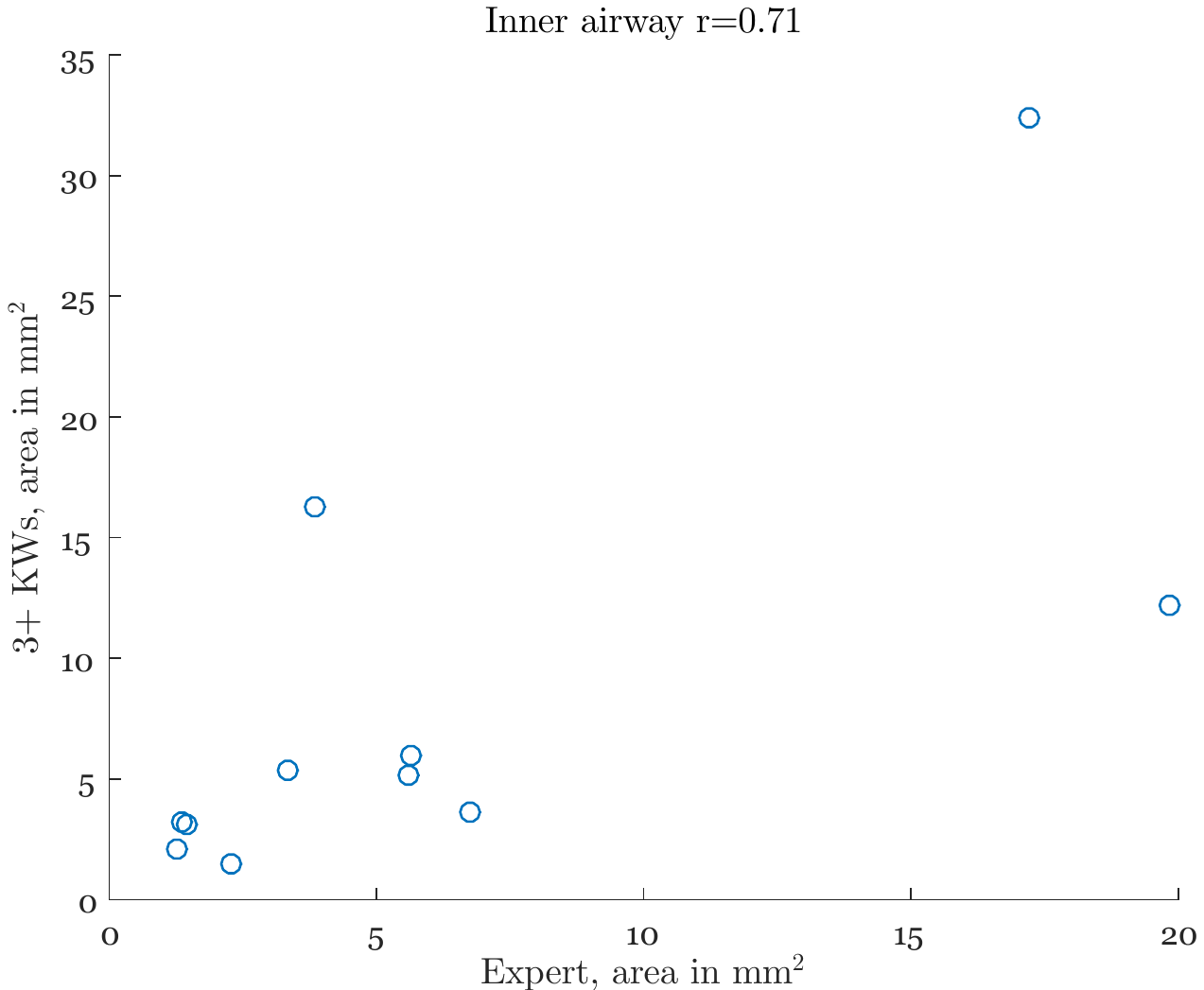}
\includegraphics[width=0.45\textwidth]{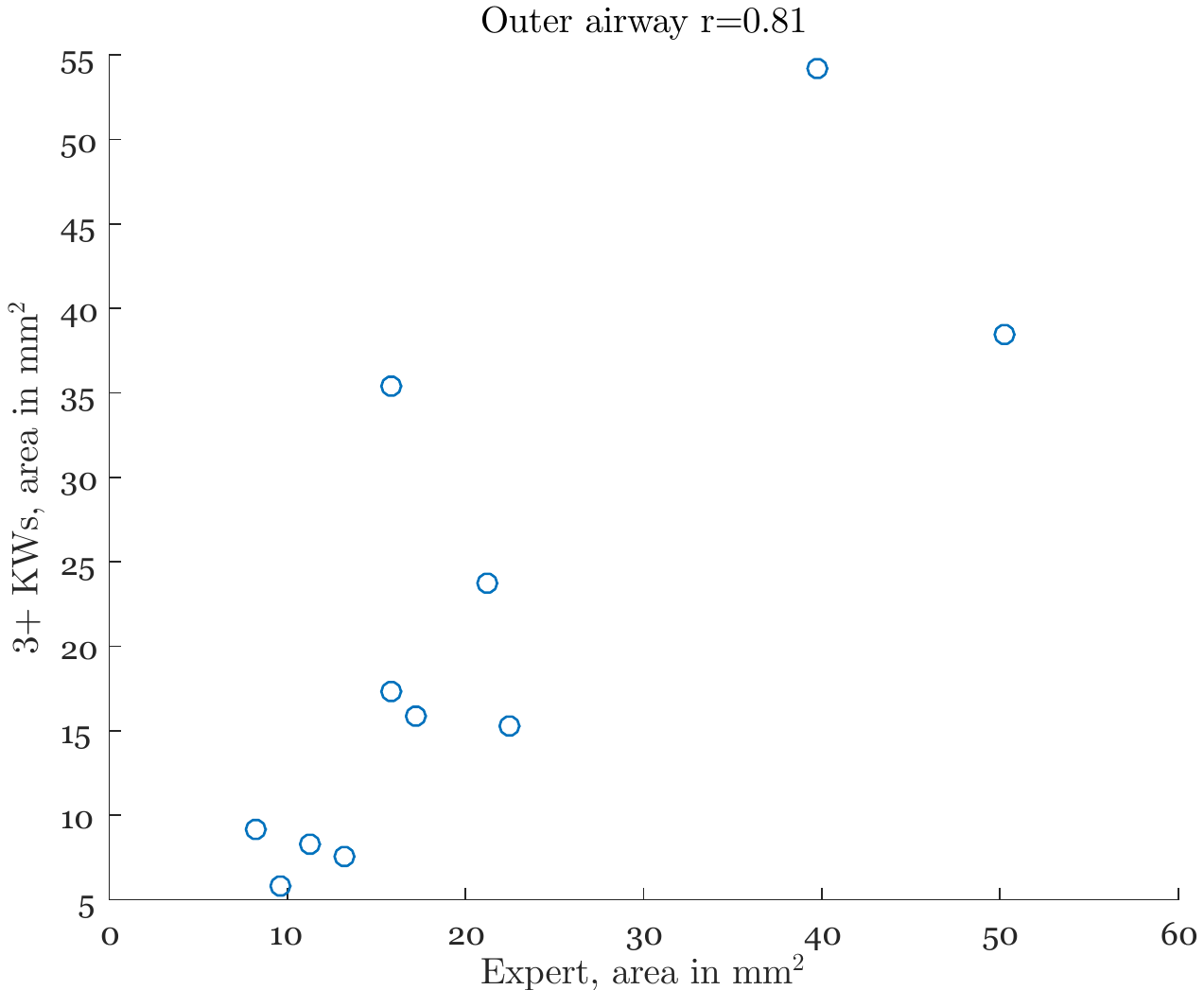}

\includegraphics[width=0.45\textwidth]{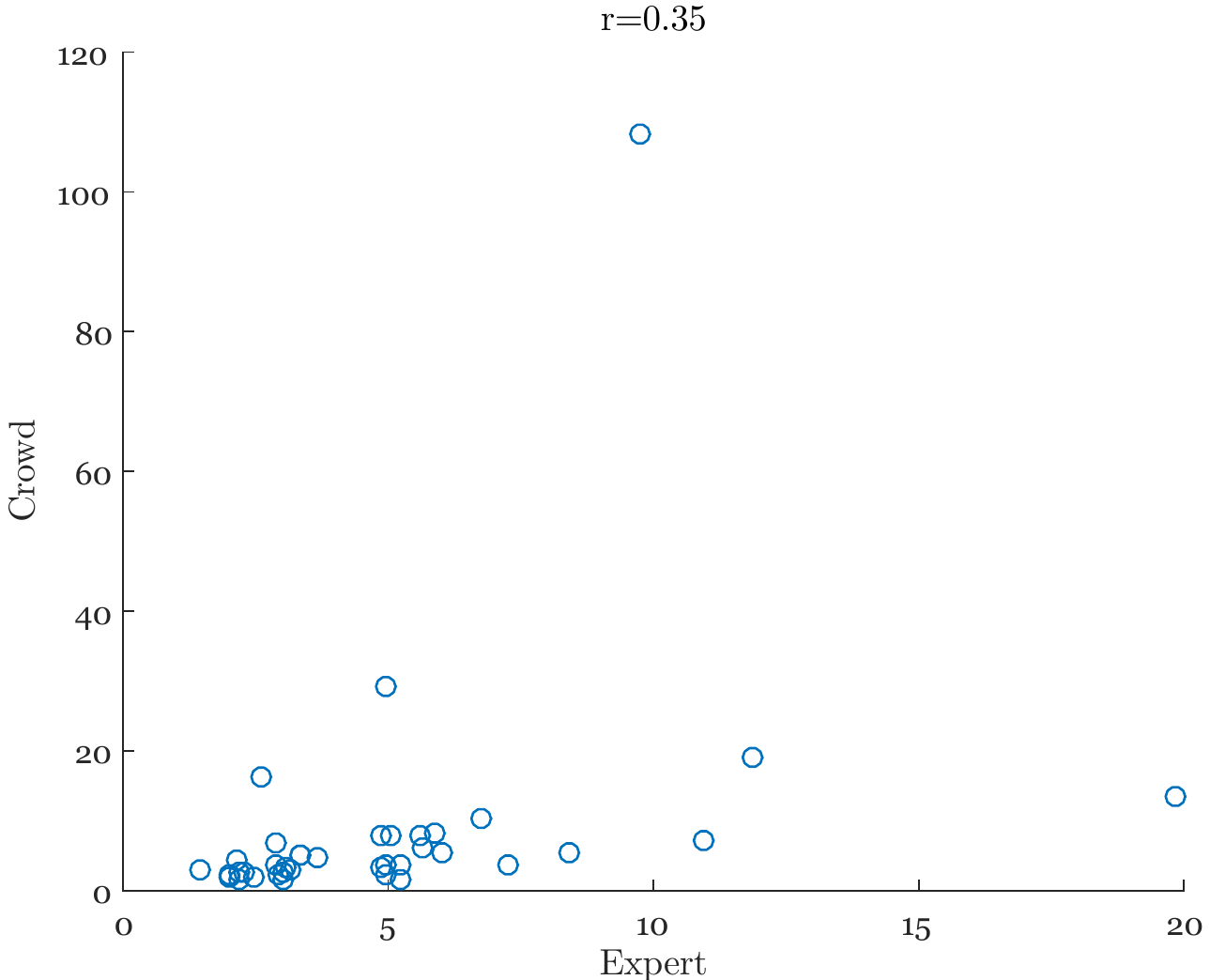}
\includegraphics[width=0.45\textwidth]{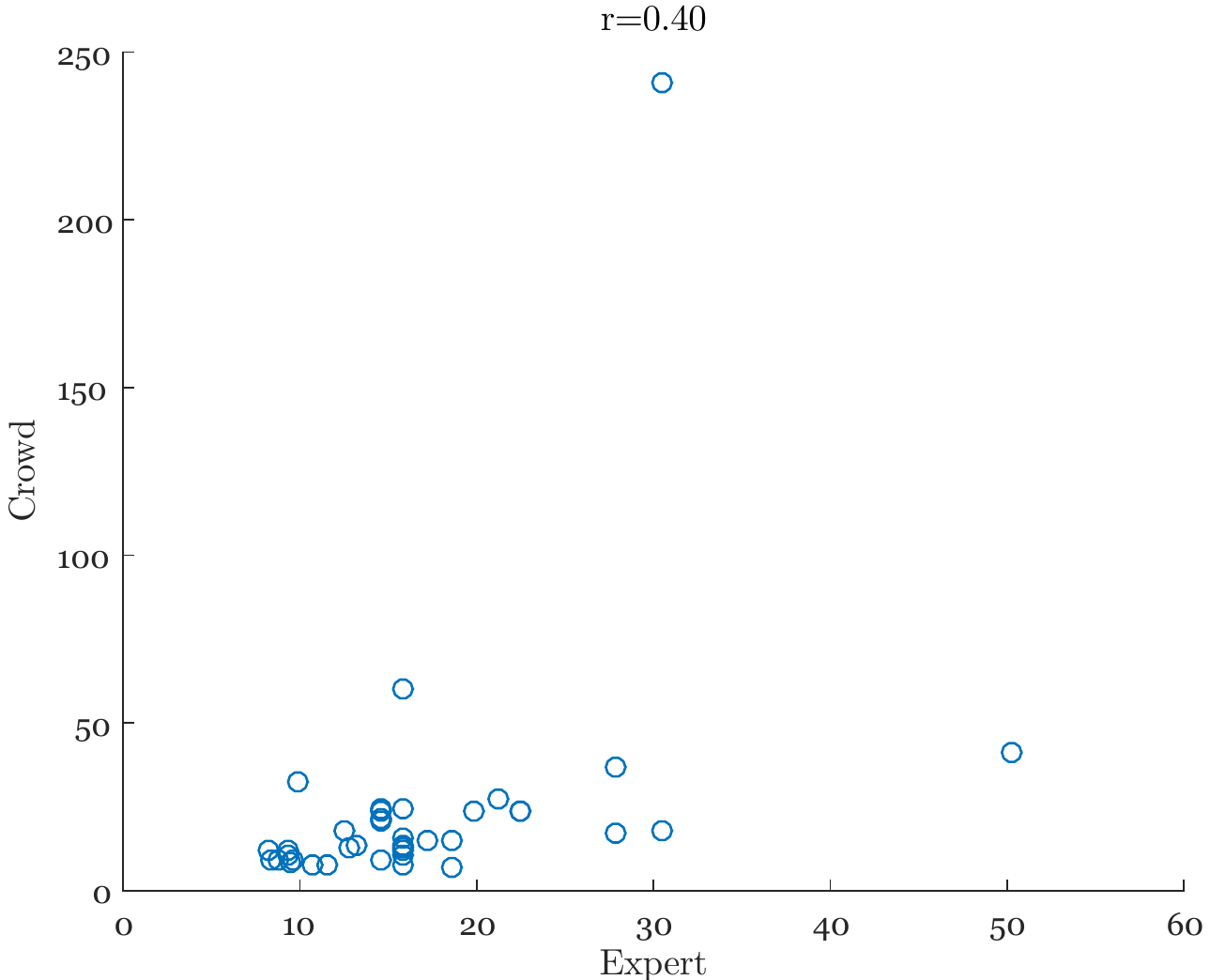}
\caption{Scatter plots of expert vs crowd (at least 3 workers) measurements of the areas. Top: original orientation, bottom: axes-parallel orientations. Left: airway lumen, right: airway wall. Left: airway lumen, right: airway wall. $r$ indicates Pearson's correlation.}
\label{fig:aggregate}
\end{figure}

Note that analysis above is performed on a per-annotation, not per-image basis. By aggregating the annotations obtained per image, we can get better estimates of the measurements from the crowd. In Fig.~\ref{fig:aggregate} we show the median areas for the images for which at least three workers produced usable annotations. The correlations are now medium to high for both types of orientations, although the sample size is lower, because for many images there were too few usable annotations. This motivates collecting more annotations per image in the future.


\section{Discussion}\label{sec:discussion}

Our results show that untrained KWs are able to interpret the CT images and attempt to annotate airways in the images. However, many KWs did not follow the instructions, resulting in unusable annotations. For example, in 244 out of 900 annotations the workers did attempt create an annotation, but only placed a single ellipse in the image.

The usable annotations show medium to high correlations with expert measurements of the airways, especially if the worker annotations are aggregated. The results are not convincing enough to say that the workers can annotate the airways as well as experts (as more analysis is needed to test such claims), but the collected annotations could already be useful for training machine learning algorithms. Overall we feel that the results encourage further investigation. The next step is to collect annotations for all 24 subjects in the cohort, after a number of changes we describe below.

Based on our results, the next logical step is to increase the amount of usable annotations per image. There are several ways in which this can be achieved. One possibility is to improve the interface, for example by only accepting annotations that contain two ellipses. Alternatively, we could include a tutorial, showing workers step by step how to create the annotations. However, both of these options require custom-made adjustments to the interface, which is costly / time-consuming for novice users of MTurk such as ourselves.

In the short term, more feasible solutions for us are to simplify the instructions, increase the number of collected annotations per image to 20 (20 is also the choice in other crowdsourcing literature~\cite{nguyen2012distributed,mitry2015crowdsourcing}), and to improve the postprocessing of the annotations. Here we used very simple rules to filter and aggregate the annotations with reasonable results. An alternative would be to use unsupervised outlier detection, or train a supervised classifier to detect outliers. Such a classifier could be based only on the characteristics of the annotations (such as size of the ellipse), or could also include characteristics of the image.

If our future research demonstrates that the crowd can reliably annotate airways, we will need to address the question of localizing the airways, and of using the annotations in machine learning algorithms. For localizing airways, we could show larger slices, and ask the KWs to click all locations where airways are visible. Such clicks can then be used to learn to recognize good voxel positions, at which airway measurements can be collected. Alternatively, we could use the already collected annotations (both usable and unusable) to learn the appearance of ``annotatable'' slice, bypassing the localizaton step.

Overall our first experiences with crowdsourcing are positive, but also teach us a number of important lessons: (i) there is more to setting up a crowdsourcing task than we thought, and (ii) the task itself needs to be simpler than we thought. With regard to setting up the task, a challenge was to make a choice between different annotation tools, and how such tools might inflence the results. With regard to the task itself, the number and the wording of instructions are likely to affect how well the instructions will be carried out.

For both the annotation interface and the instructions, it would be interesting to investigate how exactly different choices influence the final results. However, this ``parameter space'' is too large, and it is not feasible to explore it. This calls for more ``rules-of-thumb'' when designing large-scale data annotation tasks, as well as more interaction between researchers in medical image analysis, and researchers in fields where crowdsourcing is a more established technique.

\section{Conclusions}\label{sec:conclusion}

We presented our early experiences with setting up a crowdsourcing task for measuring airways in chest CT images. Our results show that the KWs were able to interpret the images, but that the instructions were too complex, leading to many unusable annotations. For the usable annotations, quantitative results show medium to high correlations with expert measurements of the airways, especially if measurements of the KWs are aggregated. Our results are encouraging, we therefore intend to continue this research direction, by simplifying the instructions and collecting more annotations for an in-depth analysis. As beginner users of crowdsourcing, we describe several challenges we encountered during this research, and we hope our experiences will help other
researchers in medical image analysis considering crowdsourcing for annotating their data.

\section*{Acknowledgements}

This research was partially funded by the research project ``Transfer learning in biomedical image analysis'' which is financed by the Netherlands Organization for Scientific Research (NWO) grant no. 639.022.010. We gratefully acknowledge Dr. Daniel Kondermann of Heidelberg University for his help with the crowdsourcing tasks.

\bibliographystyle{splncs}
\bibliography{refs}	

\end{document}